\documentclass{article}

\usepackage{arxiv}
\usepackage[utf8]{inputenc} 
\usepackage[T1]{fontenc}    
\usepackage{hyperref}       
\usepackage{url}            
\usepackage{booktabs}       
\usepackage{amsfonts}       
\usepackage{nicefrac}       
\usepackage{microtype}      
\usepackage{float}
\usepackage{xcolor}
\usepackage{graphicx}
\usepackage[ruled,vlined]{algorithm2e}
\usepackage{caption}
\usepackage{subcaption}
\usepackage{multicol}
\usepackage{booktabs}
\usepackage{adjustbox}

\title{Learning Machines from Simulation to Real World 
}

\author{%
  Tomer Iwan \\
  Vrije Universiteit Amsterdam\\
  Amsterdam, the Netherlands \\
  \texttt{t.iwan@student.vu.nl} \\
   \And
   Oktay Kavi \\
   Vrije Universiteit Amsterdam \\
   Amsterdam, the Netherlands \\
   \texttt{o.kavi@student.vu.nl} \\
  \AND
  Erkin Yildirim \\
  Vrije Universiteit Amsterdam \\
  Amsterdam, the Netherlands \\
  \texttt{e.s.h.yildirim@student.vu.nl} \\
}

\begin{document}

\maketitle
\begin{abstract}
Learning Machines is developing a flexible, cross-industry, advanced analytics platform, targeted during stealth-stage at a limited number of specific vertical applications. 
In this paper, we aim to integrate a general machine system to learn a variant of tasks from simulation to real world. In such a machine system, it involves real-time robot vision, sensor fusion, and learning algorithms (reinforcement learning). To this end, we demonstrate the general machine system on three fundamental tasks including obstacle avoidance, foraging, and predator-prey robot. The proposed solutions are implemented on Robobo robots with mobile device (smartphone with camera) as interface and built-in infrared (IR) sensors. The agent is trained in a virtual environment. In order to assess its performance, the learned agent is tested in the virtual environment and reproduce the same results in a real environment. The results show that the reinforcement learning algorithm can be reliably used for a variety of tasks in unknown environments.

\end{abstract}

\section{Introduction}

Learning machines are agents that make use of sensors, controllers, and actuators in an environment to solve a specific problem. In such machines, the mapping between sensors and actuators is not written manually. Instead, a learning algorithm is used to teach the machine what action to perform in certain states by trial and error. This is typically done by using machine learning algorithms \cite{friedberg1958learning}. General goals are that the learning machines try to continuously improve behavioural performance. 

\par The term "Learning Machines" is often mistaken for "Machine Learning". However, the difference between the two is that machine learning targets automated knowledge discovery, while learning machines is about the concept of machines learning and reprogramming themselves. Besides that, learning machines is still an unclear definition of a rather fragmented expertise. 

\par This paper poses multiple tasks that will be attempted to be solved utilizing learning machines. The main problem that stands for all three components goes as follows: \textit{to equip a given robot with learning abilities and experimentally test its performance in a real environment}.

In this paper, we apply learning machines to the three fundamental tasks as follows:
\begin{itemize}
    \item Obstacle avoidance
    \item Foraging task
    \item A predator chasing a prey
\end{itemize}

Due to the relatively simple tasks and the short time frame, two reinforcement learning algorithms, SARSA and Q-Learning were considered. Based on the ability to learn off-policy, Q-Learning was chosen, since the Q-table can be transferred from the simulation environment to the real environment.

We believe this approach is appropriate, because the implementation of simulated training into the real world is a rather new concept, which we chose to treat with the security of a well-known algorithm to us. Due to the fact that the reality gap is something that must be dealt with after successfully training in simulation, we believe and expect that the process to overcome this gap will be very time consuming \cite{jakobi1995noise}. Further motivations about the specific choice of Q-learning will be justified in the next section.

\section{Related work}
\label{related}

While learning behaviours for different task on robots, e.g., predator-prey robot problem \cite{lan2019evolutionary,lan2019simulated}, directed locomotion on evolvable modular robot \cite{lan2018directed,lan2020learning} can be already hard, it is a considerably challenge task to learn agents for various tasks, called learning machines in this paper.

We select Q-learning \cite{Sutton1998} as the learning algorithm based on the work by Barto \& Sutton about the pros and cons of Q-Learning and SARSA. 
A reason for this choice is based on their explanation of how both learning algorithms behave with regards to loss. During our simulated experiments, there is no risk of loss if the learning algorithm fails a learning trial, which eliminates a disadvantage in the greedy Q-learning algorithm. Opposed to this stands SARSA, which aims to find solutions by making as little errors as possible, possibly taking longer to learn than Q-learning.

Bing-Qiang et al. \cite{huang2005reinforcement} are faced with a similar obstacle avoidance problem. In their paper, they explain why reinforcement learning, specifically Q-learning, is a powerful tool for obstacle avoidance and tasks built upon obstacle avoidance. For tasks where the environment is unpredictable and unknown, a learning algorithm such as Q-learning is ideal. Besides the advantage in unknown environments, Q-learning is a simple, but powerful tool alongside its controller (Q-table), to store values of previous actions. Another advantage given by Q-learning is that it works off-policy, which means no online learning is required. In this way, a learned model can simply be transferred.

\section{Methodology}

Identical procedures of methodology were applied to all three tasks, with obvious required deviations per task. Below, the main methodology is explained.

\subsection{The proposed approach}

The proposed approach for each task followed the same method. First, all possible states for each task were evaluated and worked into rows of the controller table to the learning algorithm. This is called the state space. The controller will be treated in depth in the next section. After evaluating the possible states, each possible action was evaluated as well and put into columns of the controller table. This is called the action space. After forming the proper spaces, the reward function was defined for each state, which can be positive or negative.

After each action, the controller is updated, values are kept in order to improve learning. The pseudocode for the updating of values goes as follows \cite{Sutton1998}:

\begin{algorithm}[!htbp]
\caption{Q-learning (off-policy TD control)}
\SetAlgoLined
 initialize $Q(s,a)$, $\forall s \in S$, $a \in A(s)$, arbitrarily, and $Q$(terminal-state, $\bullet$) = 0\;
\For{each episode}
 {
  initialize S\;
  \For{each step of episode}
  {
      Choose A from S using policy derived from Q\;
      Take action A, observe R, S'\;
      $Q(S,A) \gets (1- \alpha) \cdot Q(S,A) + \alpha [R + \gamma \cdot max_a Q(S', a) - Q(S,A)]$\;
      $S \gets S'$\;
  }
 }
 \end{algorithm}

The $\alpha$ denotes the learning rate, $\gamma$ denotes the discount rate, R denotes the reward, $S$ denotes the current state, and $S'$ signifies the next state.

\subsection{The controller representation}

In each task, the controller that is used alongside the learning algorithm is a Q-table. This table contains the state space mapped against the action space. The values in this table represent the Q-values based on the acquired reward when being in state S and performing action A.

\subsection{The learning algorithm}
After each performed action made by the learning machine, it receives feedback called "reward". If we want to positively reinforce certain behaviour in a certain state, this reward is positive. If we want to negatively reinforce certain behaviour, it receives a negative reward. The learning algorithm makes sure that for each action, the action in the previous state has its reward updated depending on whether it was desired or not. This happens by means of the following Bellman's equation \cite{sundar1997optimal}:

\begin{equation}
Q(s,a) = (1-\alpha) \cdot Q(s,a) + \alpha \cdot ( r(s,a) + \gamma \cdot max_a( Q'(s',a)) - Q(s,a))
\end{equation}

in which max chooses the maximum expected reward for the next state, s'. Q(s,a) is the value for the chosen action in the current state, which will then be updated in the controller (Q-table).

\section{Experiments}

The Obstacle Avoidance task requires a Robobo robot to explore an unknown environment in which it is should avoid obstacles and walls. The robot should learn how to detect an obstacle and how to avoid collision with the obstacle using the IR-sensors. In the end, it should be able to navigate through the environment while successfully avoiding any walls and obstacles and not repeatedly covering the same part of the environment.

In the Foraging task, a Robobo robot should be able to navigate its way through an unknown environment, like in the previous task, but should focus on finding and "eating food" using the IR-sensors and an external camera. In this case, food is represented as a green object and eating implies that Robobo comes in contact with the food. While doing so, Robobo should also avoid minor obstacles and walls.

In the Prey-Predator task, Robobo should be able to catch its prey. The robot should learn how to spot its prey (another Robobo), close the distance and follow the prey using the IR-sensors and an external camera. A red colour is used for the prey to distinguish it from the environment. 

The parameters used for all tasks can be found below in Table \ref{tab:table-name}.

\begin{table}[!htbp]
\centering
\renewcommand{\arraystretch}{1.2}
    \begin{tabular}{ p{2cm}p{2cm} p{2cm} p{2cm}}
    \hline
    --- & Task 1 & Task 2 & Task 3 \\
    \hline
    Learning Rate & 0.1 & 0.1 & 0.1\\
    Discount &   0.9 & 0.9 & 0.9\\
    Epsilon & 0.6 - 0.1 & 1.0 - 0.1 & 1.0 - 0.1 \\
    Epochs    & 5 & 10 & 10\\
    Steps & 1000 & 1000 & 500\\
    \hline                                                                 
    \end{tabular}
\caption{\label{tab:table-name}Main parameters of Q-learning for all tasks.}
\end{table}

The simulations started with an empty Q-table (zeroes) at the first Epoch. When in a particular state, Robobo was rewarded or punished for certain actions. After each epoch, the epsilon value decreases by 0.1, promoting exploration in the beginning and exploitation when enough exploration has been done.

\par  For most cases, perceiving targets in the environment using vision is crucial for various tasks in robotics.
However, the limitations of the computing hardware make this a challenging task.
In \cite{lan2019evolving,lan2018ICARCV}, G. Lan et al. proposed the methods for real-time object recognition on small robots with low-performance computing hardware, where targets with color paper are recognized with OpenCV. We therefore use the similar solution to recognize targets in this paper due to we use the similar small robots with low-performance computing hardware.
Image processing to allow Robobo to spot food objects or the prey was handled with OpenCV, a python image processing library. OpenCV allows the external camera sensor to sense BLOBs (Binary Large OBjects). Target objects were classified as desired BLOBs. Depending on which part of the image the BLOB was present, Robobo should learn to move towards it. At the end of the last epoch, we are left with a fully explored Q-table.\par

\begin{figure}[!htbp]
    \centering
    \includegraphics[width=0.8\columnwidth]{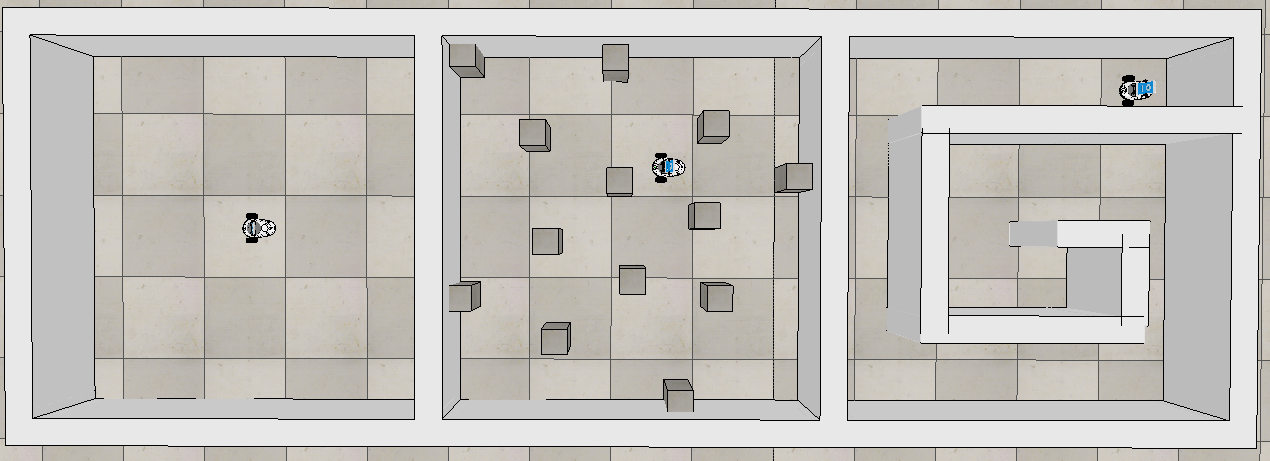}
    \caption{The three maps: (i) no obstacles, (ii) obstacles and (iii) maze.}
    \label{fig:OBSMAPS}
\end{figure}

For the obstacle avoidance task, the state and action space can be found in Appendix A \ref{task_1_SR} and \ref{task_1_A}.
As there was no goal state for the obstacle avoidance task, the number of steps was chosen arbitrarily as the terminating state.
The performance measures for the obstacle avoidance task were the cumulative reward, the number of crashes, and the number of bad decisions made by the robot. 
Chosen was to reward whenever nothing was detected, and to punish whenever any IR-sensor detected an object. The environment had 3 sub environments: Just walls, with obstacles, and a maze. The environments can be seen in Figure \ref{fig:OBSMAPS}. We decided to initialize our Q-table in the first environment for 5 epochs, starting at epsilon 0.6, followed by training in the second sub environment, again starting with epsilon 0.6. And finally, training in the last sub environment for 5 epochs. The Q-table was transferred from one sub environment to the other. In total, 15 epochs were performed to acquire the final Q-table.

{}

For the foraging task, the state and action space can be found in Appendix A \ref{task_2_SR} and \ref{task_2_A}.
This task had a goal state, which was reached when all green food objects in the simulated arena were eaten. At the beginning of each epoch, the location of food objects in the simulation arena was randomized. This is to prevent overfitting to a limited set of data points. \par
A model for food foraging was determined by having the simulated Robobo learn to consume all food as fast as possible. The performance measure here is the average steps per food and the cumulative reward.
Chosen was to reward whenever the camera spotted a food object. Spotting the food object closer was rewarded more than spotting it in the distance. A punishment was given whenever nothing was detected, or the IR-sensors detected an obstacle. A higher punishment was given whenever an obstacle was detected, since being stuck in an obstacle could greatly impact the time it takes to collect all the food.\par

For the prey-predator task, the state and action space can be found in Appendix A \ref{task_3_SR} and \ref{task_3_A}.
The aim here is to catch the prey. However, since it is difficult to define when the predator has caught the prey using only the camera which faces one direction, and the IR sensors, we decided to have the number of steps as the terminal state. 
Chosen was to reward whenever the camera spotted the prey. Spotting the prey closely and in the center was rewarded the highest, followed by spotting the prey closely on the left or the right side, and finally, the lowest reward was given whenever the prey was spotted in the distance. Any other state was punished with the same negative reward. \par

\subsection{Task 1: Obstacle avoidance}

\begin{figure}[!htbp]
     \centering
     \begin{subfigure}[b]{0.49\textwidth}
         \centering
         \includegraphics[width=\textwidth]{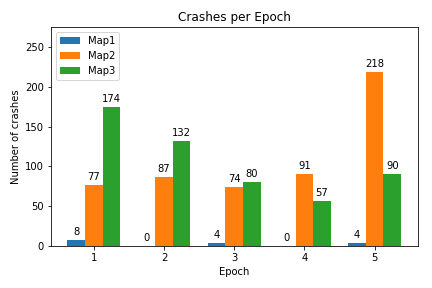}
         \caption{Number of crashes per epoch for each of the three maps.}
         \label{1a}
     \end{subfigure}
     \hfill
     \begin{subfigure}[b]{0.49\textwidth}
         \centering
         \includegraphics[width=\textwidth]{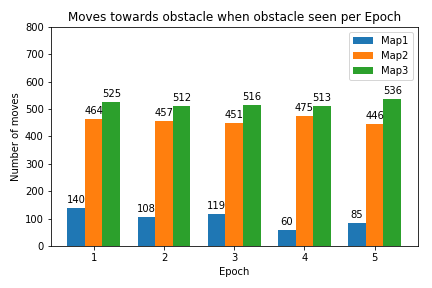}
         \caption{Number of wrong actions per epoch for each of the three maps.}
         \label{1b}
     \end{subfigure}
     \hfill
     \begin{subfigure}[b]{0.52\textwidth}
         \centering
         \includegraphics[width=\textwidth]{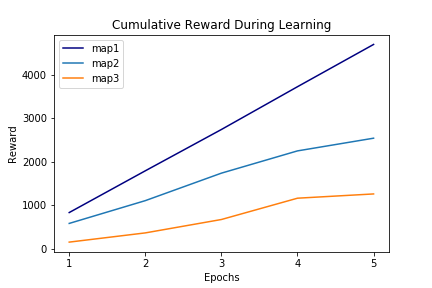}
         \caption{Cumulative reward during obstacle avoidance task.}
         \label{1c}
     \end{subfigure}
        \caption{Results of obstacle avoidance task.}
        \label{fig:1}
\end{figure}

A general metric for tasks concerning Q-learning is that of cumulative reward. In Fig. \ref{fig:1}(c) the cumulative reward over all the epochs is shown for each sub environment. The plot shows that in the first epochs a lot of negative reward is accumulated, leading to the declining trend visible. Afterwards, positive rewards make sure to bring the cumulative reward up over each epoch. 

The number of crashes in each epoch is depicted in Fig. \ref{fig:1}(a). A crash is defined as a situation in which the IR sensors have detected a value close to the max value. Observed can be that the number of crashes was very low in the first sub environment. The second sub environment had an increasing number of crashes, and the last sub environment had a decreasing number of crashes.

Finally, in Fig. \ref{fig:1}(b) we have decided to show the number of bad decisions. A bad decision in this context is when the agent turns in the direction of a wall or obstacle, instead of moving away from it. The first sub environment shows a slight decrease in bad decisions, but the other two sub environments have remained stable. 

Overall, the results were not what we expected, especially the number of crashes and the number of bad decisions. Perhaps this was due to the difficulty of the environment, as there were many obstacles close to each other in the second sub environment. As for the maze, the walls were very close to the robot, which may have made it difficult for it to navigate through the environment without crashing. Changing the speed of the actuators and the IR-sensor threshold, however might lead to a better performance in the future. 

\subsection{Task 2: Foraging task}
\begin{figure}[!htbp]
     \centering
     \begin{subfigure}[b]{0.49\textwidth}
         \centering
         \includegraphics[width=\textwidth]{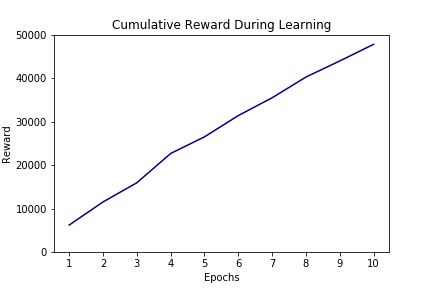}
         \caption{Cumulative reward during food foraging task.}
         \label{2a}
     \end{subfigure}
     \hfill
     \begin{subfigure}[b]{0.49\textwidth}
         \centering
         \includegraphics[width=\textwidth]{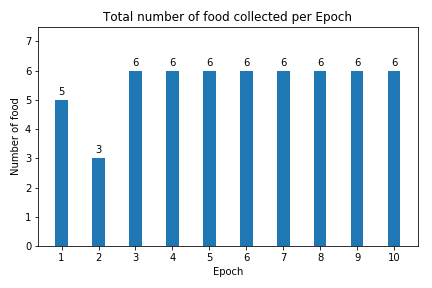}
         \caption{Number of food collected in each epoch.}
         \label{2b}
     \end{subfigure}
     \hfill
     \begin{subfigure}[b]{0.52\textwidth}
         \centering
         \includegraphics[width=\textwidth]{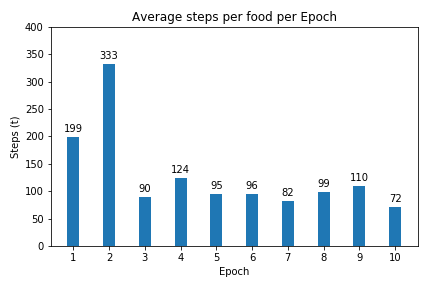}
         \caption{Average time needed to collect a food.}
         \label{2c}
     \end{subfigure}
        \caption{Results of food foraging task.}
        \label{fig:2}
\end{figure}

In this task, we have decided to once again look at the cumulative reward. The plot in Fig. \ref{fig:2}(a) shows a linear increase in the cumulative reward.

This experiment did have a goal state, namely the collection of all food items. In total there were seven items which after each epoch were put in different locations in the arena. Chosen was to terminate the epoch after collecting six out of seven, as the function responsible for relocating the food sometimes caused food to be generated outside of the arena. In Fig. \ref{fig:2}(b) we see that after learning, the agent is able to collect all food items. 

To have a closer look at the results we decided to look at the average time needed to collect all food items. It can be observed in Fig. \ref{fig:2}(c) that the average time went down after learning.

Overall, we noticed that even with randomized food locations, the robot could consistently collect all food objects in a little more than half the total steps given. 

\subsection{Task 3: A predator chasing a prey}

\begin{figure}[!htbp]
     \centering
     \begin{subfigure}[b]{0.49\textwidth}
         \centering
         \includegraphics[width=\textwidth]{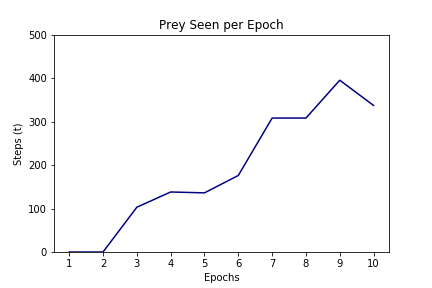}
         \caption{How many time steps the prey was seen.}
         \label{3a}
     \end{subfigure}
     \hfill
     \begin{subfigure}[b]{0.49\textwidth}
         \centering
         \includegraphics[width=\textwidth]{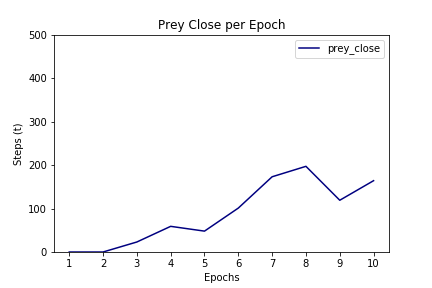}
         \caption{How many time steps the prey was seen closely.}
         \label{3b}
     \end{subfigure}
     \hfill
     \begin{subfigure}[b]{0.49\textwidth}
         \centering
         \includegraphics[width=\textwidth]{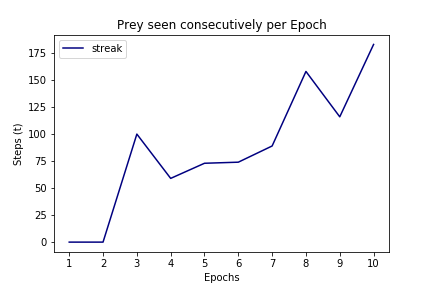}
         \caption{Prey seen consecutively per epoch}
         \label{3c}
     \end{subfigure}
     \begin{subfigure}[b]{0.49\textwidth}
         \centering
         \includegraphics[width=\textwidth]{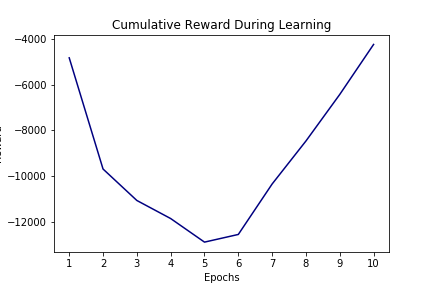}
         \caption{Cumulative reward in the predator-prey task}
         \label{preycum}
     \end{subfigure}
        \caption{Results of predator-prey task.}
        \label{fig:3}
    \centering
\end{figure}

As with the previous tasks, the performance measure of cumulative reward was taken into consideration. In Fig. \ref{fig:3}(d), we can see a negative trend in the first half of the experiment, followed by a positive trend in the last five epochs. This can be mainly explained due to the change in epsilon value after each epoch. In the beginning, epsilon starts high to promote exploring, whilst near the end the epsilon value is low to promote exploiting what was learned during training. 

Next, in Fig. \ref{fig:3}(a) we observe how often prey was spotted with the camera by our predator agent. A positive trend is seen as the number of times during the run increases with each epoch.

To get an indication of whether the predator has seen the prey from close or far, we chose to differentiate between the two in the state space. In Fig. \ref{fig:3}(b), we can observe that this performance measure has also increased over each epoch.

Finally, in Fig. \ref{fig:3}(c) we can observe that during learning the agent was able to follow the prey for longer periods of time than at the start of learning.  This observation is in-line with our expectation that after learning the Robobo will be able to follow the prey over longer periods of time during a simulation run.

On the difference between learning behaviour for a real machine and a simulated one, we can discuss that the reality gap should be taken into consideration. In our experiments, the reality gap was noticeable in both sensors and behaviour. The blob detection for specific colours was different in simulation from reality. Also, the hardware differences were noticeable as the actuators were not perfectly even just as in the simulation. 

Overall, we can see that the robot shows improved behaviour over time. We were especially interested in whether the robot could learn to follow its prey, and the results have shown that the robot learned to follow its prey for a longer period.

\section{Conclusions}

In this paper, our aim was \emph{to equip a given robot with learning abilities and experimentally test its performance in a real environment}. Using Q-Learning in a simulated environment, we attempted to provide a solution to three tasks by defining the state space, action space and a reward function for each task. 

In the first task of obstacle avoidance, our proposed solution does not lead to an increase in performance at all times. Especially for the number of bad decisions, where no decrease was seen after learning in 2 of the three sub environments. As for the number of crashes, these also remained quite high compared to our expectations.

In the food foraging task, we conclude that the performance of the agent indeed improved after the learning process, as the time needed to complete the task decreased. 

In the final task, the main goal of the learning agent was to catch its prey. In addition, we gave it the sub-goal of accomplishing the task whilst following the prey. This also was considered a success as not only was out predator robot able to catch the prey, it also showed improvement in the time it consecutively followed the prey, before eventually catching it. 

Regarding the experience of simulated robots versus real-life robots and the reality gap, we have learned that agent and environmental differences posed the biggest obstacle. The simulated learning agent was consistent in its actuator performance (speed and power), as opposed to actuator performance of the real-life robot, which was often impaired on one side, causing a skew in movement. 

As for the environment, the simulated environment was naturally unchanged for each simulation, as opposed to real-life. The ambience affected the performance of image processing greatly, and the simulated environment did not exactly scale 1:1 with the real-life environment. These are parameters that must be kept in check in order to minimize the effects of the reality gap. In addition, we noticed that the perception with IR sensors is not robust enough. In future work, we aim to combine the sensing technology \cite{lan2016development,lan2016bayesian,liu2016convolution,hanbo2019online} to improve the robustness.

To conclude, we have shown in this paper that the simple reinforcement Q-learning algorithm was more than sufficient to carry out the tasks at hand.

\bibliographystyle{plain}
\bibliography{main}
\appendix
\section{Appendix}

\begin{table}[!htbp]
    \begin{subtable}[h]{0.45\textwidth}
        \centering
        \begin{tabular}{|l|l|}
                \hline
                States:                        & Reward: \\ \hline
                Object Front                   & -1      \\ \hline
                Object Left                    & -1      \\ \hline
                Object Right                   & -1      \\ \hline
                Object Back Left               & -1      \\ \hline
                Object Back Right              & -1      \\ \hline
                Object Back Left \& Back Right & -1      \\ \hline
                Nothing Detected               & 1       \\ \hline
            \end{tabular}
       \caption{States \& Rewards}
       \label{task_1_SR}
    \end{subtable}
    \hfill
    \begin{subtable}[h]{0.45\textwidth}
        \centering
            \begin{tabular}{|l|}
                \hline
                Actions:   \\ \hline
                Forwards   \\ \hline
                Up Left    \\ \hline
                Up Right   \\ \hline
                Back Left  \\ \hline
                Back Right \\ \hline
                Backwards  \\ \hline
            \end{tabular}
        \caption{Actions}
        \label{task_1_A}
     \end{subtable}
     \caption{States, Rewards and Actions of Task 1}
\end{table}

\begin{table}[!htbp]
    \begin{subtable}[b]{0.45\textwidth}
        \centering
        \begin{tabular}{|l|l|}
\hline
States:                              & Reward: \\ \hline
Target Far Left                      & 5       \\ \hline
Target Far Center                    & 5       \\ \hline
Target Far Right                     & 5       \\ \hline
Target Close Left                    & 10      \\ \hline
Target Close Center                  & 10      \\ \hline
Target Close Right                   & 10      \\ \hline
Object Front                         & -10     \\ \hline
Object Left                          & -10     \\ \hline
Object Right                         & -10     \\ \hline
Nothing Detected but last seen Left  & -1      \\ \hline
Nothing Detected but last seen Right & -1      \\ \hline
Nothing Detected                     & -5      \\ \hline
\end{tabular}
       \caption{States \& Rewards}
       \label{task_2_SR}
    \end{subtable}
    \hfill
    \begin{subtable}[b]{0.45\textwidth}
        \centering
            \begin{tabular}{|l|}
                \hline
                Actions:   \\ \hline
                Forwards   \\ \hline
                Up Left    \\ \hline
                Up Right   \\ \hline
                Back Left  \\ \hline
                Back Right \\ \hline
                Backwards  \\ \hline
            \end{tabular}
        \caption{Actions}
        \label{task_2_A}
     \end{subtable}
     \caption{States, Rewards and Actions of Task 2}
\end{table}

\begin{table}[!htbp]
    \begin{subtable}[b]{0.45\textwidth}
        \centering
        \begin{tabular}{|l|l|}
\hline
States:                              & Reward: \\ \hline
Target Far Left                      & 5       \\ \hline
Target Far Center                    & 5       \\ \hline
Target Far Right                     & 5       \\ \hline
Target Close Left                    & 10      \\ \hline
Target Close Center                  & 20      \\ \hline
Target Close Right                   & 10      \\ \hline
Object Front                         & -10     \\ \hline
Object Left                          & -10     \\ \hline
Object Right                         & -10     \\ \hline
Nothing Detected but last seen Left  & -10      \\ \hline
Nothing Detected but last seen Right & -10      \\ \hline
Nothing Detected                     & -10      \\ \hline
\end{tabular}
       \caption{States \& Rewards}
       \label{task_3_SR}
    \end{subtable}
    \hfill
    \begin{subtable}[b]{0.45\textwidth}
        \centering
            \begin{tabular}{|l|}
                \hline
                Actions:   \\ \hline
                Forwards   \\ \hline
                Up Left    \\ \hline
                Up Right   \\ \hline
                Backwards  \\ \hline
                Turn Around Axis (Clockwise) \\
                or Forwards (50\% chance) \\ \hline
                Turn Around Axis (Counter-Clockwise) \\ or Forwards (50\% chance)  \\ \hline
            \end{tabular}
        \caption{Actions}
        \label{task_3_A}
     \end{subtable}
     \caption{States, Rewards and Actions of Task 3}
\end{table}
\end{document}